\documentclass[numeric]{kinamind}
\usepackage{float}

\newcommand{\cframework}{C4}
\newcommand{\csetfull}{C4 Evaluation Set}
\newcommand{\cset}{C4-Eval}

\title{Can MLLMs Decode the Creative Leap? Introducing \cframework{} for Cross-Concept Understanding}
\shorttitle{C4 for Cross-Concept Understanding}
\author[1,2]{Ming Wang\orcid{0000-0001-8406-5677}}
\author[3]{Yuqing Zhang\orcid{0009-0006-2834-2864}}
\author[1]{Tingna Xie}
\author[3]{Xiangju Li}
\author[1]{Xiaocui Yang\orcid{0000-0001-5352-8579}}
\author[1]{Daling Wang\orcid{0000-0003-1340-0778}}
\author[1]{Shi Feng\orcid{0000-0002-2846-7652}}
\author[1]{Yifei Zhang\orcid{0000-0003-0854-2966}}
\affil[1]{School of Computer Science and Engineering, Northeastern University, Liaoning, China}
\affil[2]{School of Computing and Information Systems, Singapore Management University, Singapore}
\affil[3]{School of Computer Science and Engineering, Shandong University of Science and Technology, Shandong, China}
\venue{arXiv preprint}
\hypersetup{pdfauthor={Ming Wang, Yuqing Zhang, Tingna Xie, Xiangju Li, Xiaocui Yang, Daling Wang, Shi Feng, Yifei Zhang}}

\begin{document}

\maketitle

\begin{abstract}
Creative capabilities of MLLMs matter in design, communication, education, and human--AI collaboration, yet remain difficult to evaluate because explicit targets and reward signals are scarce compared with accuracy-oriented tasks. Cross-concept understanding is a core cognitive capacity underlying receptive creativity. It enables a perceiver to recover intended meaning from non-obvious but meaningful conceptual relations. We operationalize item construction as cross-concept encoding and model inference as cross-concept decoding. We introduce \cframework{}, a cognition-inspired evaluation framework for Chengyu (Chinese idiom)-based Cross-Concept Creativity. Its encoding component maps target slots to imageable substitute concepts along bridge paths in a manually annotated and third-party-reviewed cross-concept network, enabling batch generation with explicit structure, difficulty indexed by bridge count and depth, and exact answers. Using this framework, we instantiate the \csetfull{} (\cset{}), comprising 184 synthetic items and 37 human-created cross-concept chengyu figures collected from online sources. We manually construct and review cross-concept relations, bridge paths, and reasoning processes for the collected figures. Each \cset{} item is instantiated in five task settings, yielding 884 primary answer-recovery cases. Across ten evaluated MLLMs, the strongest closed models reach 50.7\% and 48.0\% primary accuracy, while open-source models remain substantially lower. Candidate constraints improve accuracy sharply, but bridge hints and explanation requests provide only modest gains. These results expose a substantial gap in how current MLLMs decode creatively encoded meaning through cross-concept relations. Release details are provided in the appendix.
\end{abstract}

\section{Introduction}

Creative capabilities are increasingly important as multimodal large language models (MLLMs) participate in design, education, communication, and human--AI collaboration \cite{yin2024survey,zhou2024generative,doshi2024generative}. Evaluating these capabilities, however, remains difficult \cite{huang2025causality,chakrabarty2024art}. Many accuracy-oriented tasks provide explicit targets and direct reward signals, while creative tasks involve novelty, appropriateness, and potentially multiple valid interpretations \cite{runco2012standard,huang2025causality}. These properties make modeling objectives harder to specify and scalable evaluation harder to construct, leaving creative understanding less systematically studied in current multimodal benchmarks \cite{huang2025causality,chakrabarty2024art}.
Cognitive psychology provides a principled foundation for evaluating creative understanding. Creativity combines originality with effectiveness \cite{runco2012standard}. Converging cognitive accounts place cross-concept connection at the center of creative cognition. Creative cognition emphasizes generative recombination \cite{finke1992creative}, associative theory links creativity to remote conceptual connections \cite{mednick1962associative}, structure mapping formalizes analogical alignment \cite{gentner1983structure}, and conceptual integration explains how meaning emerges from partially shared input spaces \cite{fauconnier1998conceptual}. Together, these accounts identify the ability to connect concepts that are not obviously adjacent and traverse the resulting interpretable relation as a core cognitive capacity underlying receptive creativity. We refer to this capacity as \textbf{cross-concept understanding} and to each recoverable link as a cross-concept relation. A red fruit and a physicist share no surface similarity, yet an apple links them through a well-known anecdote. Recovering the apple from such displaced clues illustrates the cognitive operation that supports creative interpretation.

\begin{figure}[t]
\centering
\includegraphics[width=\linewidth]{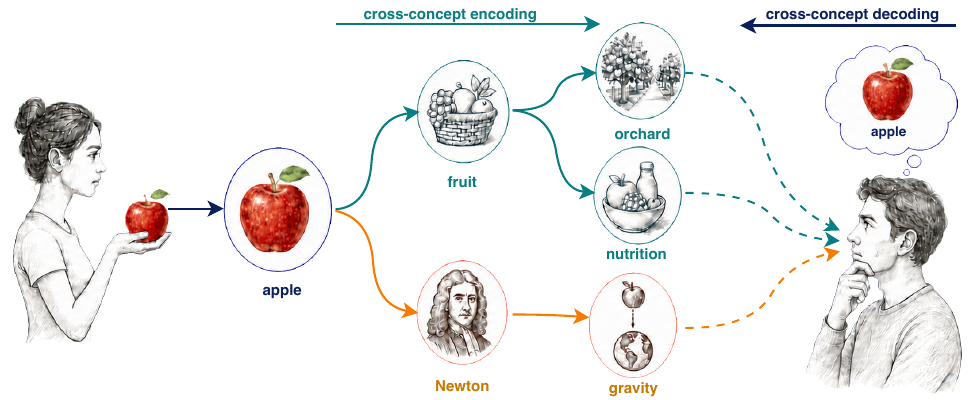}
\caption{Cross-concept encoding and decoding. A communicator encodes an apple into indirect concepts through distinct associative paths. An observer integrates the resulting clues and decodes the latent source.}
\label{fig:crossconcept-cognition}
\end{figure}

This capability admits a natural two-direction formulation, illustrated in Figure~\ref{fig:crossconcept-cognition}. We define \textbf{cross-concept encoding} as the substitution process. A communicator starts from a source concept, projects it through meaningful associative paths, and expresses it as displaced but imageable cues. We define \textbf{cross-concept decoding} as the recognition direction. An observer sees the displaced cues, hypothesizes the latent relations, and follows them in reverse to recover the source. 
Grounding this formulation requires a target space whose answers are stable and whose internal structure supports rich associations. Chinese four-character idioms, or \textbf{chengyu}, satisfy both requirements. A chengyu is a conventional expression with a fixed written form, which yields exact answers. Its characters, fragments, pronunciations, referenced objects, and cultural allusions all provide anchor points for associative substitution. This combination preserves culturally grounded creative associations while providing exact targets and inspectable paths. We use chengyu as the carrier and propose \cframework{}, a cognition-inspired evaluation framework for Chengyu-based Creative Cross-Concept understanding that makes creativity measurable at scale. 
The \cframework{} framework organizes benchmark construction and evaluation around a manually annotated chengyu-oriented cross-concept network. Two annotators independently selected concept slots from each target chengyu, associated outward from those slots, and built cross-concept bridges toward imageable landing concepts. A second round combined internal cross-checking between the two annotators with review by a third annotator, and produced the final network. On top of the network, the encoding component fixes a target chengyu, selects slots, follows bridge paths to substitute concepts, and renders the substitutes as visual clues, so item batches carry explicit latent structure, graded difficulty, and exact answers.

Using the \cframework{} framework and the reviewed network, we instantiate a concrete evaluation set, the \csetfull{} (\cset{}). \cset{} contains synthetic items at four difficulty levels, L1 through L4, indexed by bridge count and bridge depth, together with human-created figures collected from the web and manually annotated with the same cross-concept structure. 
The \cframework{} framework supports multiple task forms. Four answer-bearing settings, T1 through T4, constitute the primary evaluation, and a known-answer explanation setting, T5, is excluded from the score and reserved for analysis.
\begin{figure}[t]
\centering
\includegraphics[width=\textwidth]{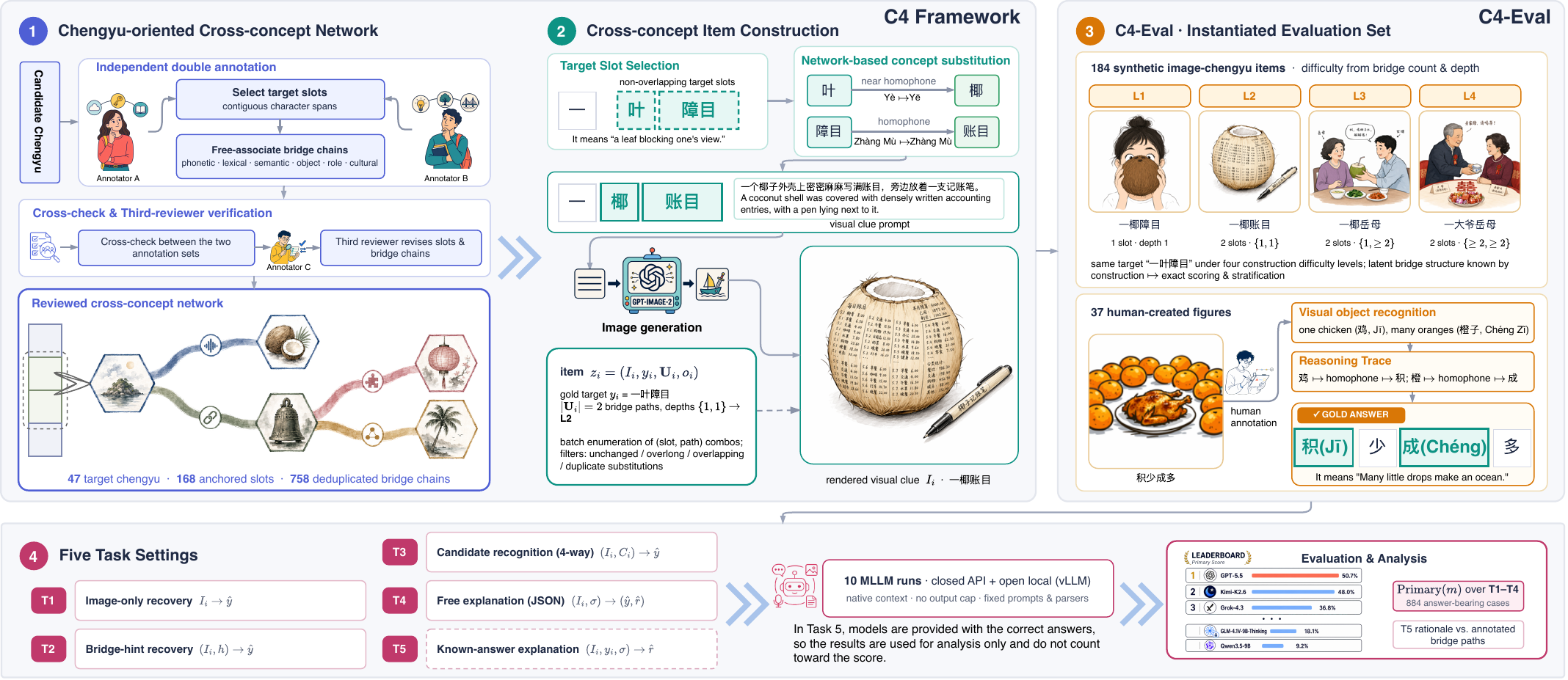}
\caption{Overview of the \cframework{} framework and \cset{}. A reviewed chengyu-oriented cross-concept network supports controlled synthetic item construction, while human-created figures receive object, reasoning-trace, and gold-answer annotations. Both sources enter five task settings and a unified MLLM evaluation, where T1--T4 determine the primary score and T5 is reserved for rationale analysis.}
\label{fig:overview}
\end{figure}
We evaluate ten closed and open multimodal models under a strict protocol. The main findings are threefold. First, the task remains far from saturated. The strongest model reaches 50.7\% primary accuracy. Second, candidate constraints add 17.3--56.0 points over open recovery, indicating that models often hold enough signal to recognize the answer in a small set but fail to decode it freely. Third, open models under native-context, no-output-cap policy lag behind the strongest closed models, even when documented reasoning wrappers are parsed through an explicit final-answer extraction policy. Together, these results reveal a substantial gap in the ability of current MLLMs to decode creatively encoded meaning.

In summary, this paper makes three contributions:
\begin{itemize}
\item We operationalize cross-concept understanding as encoding and decoding and develop the \cframework{} framework around a manually annotated and reviewed chengyu-oriented network with explicit bridge paths and difficulty levels.
\item We instantiate the framework as \cset{}, which contains 184 synthetic items and 37 web-collected figures across 84 targets and 884 primary cases with manually constructed cross-concept networks.
\item We evaluate ten MLLMs under a strict protocol, organize the analysis around explicit research questions on decoding bottlenecks and slot and bridge-depth difficulty, and systematically analyze explanation fidelity and shared failures.
\end{itemize}

\section{Related Work}

\subsection{Multimodal Reasoning Benchmarks}

Multimodal benchmarks span image-conditioned question answering, controlled and real-image compositional reasoning, external knowledge, grounded language, and image-text compositionality \cite{antol2015vqa,goyal2017vqav2,johnson2017clevr,hudson2019gqa,marino2019okvqa,suhr2019nlvr2,thrush2022winoground}. VQA and VQAv2 emphasize image-grounded answering, while CLEVR and GQA probe controlled and real-image compositional structure. MME, MM-Vet, MMMU, and CMMMU broaden coverage of perception, cognition, and domain reasoning across visual formats \cite{fu2025mme,yu2024mmvet,yue2024mmmu,zhang2024cmmmu}. REBUS targets indirect visual semantics in image-based wordplay through image recognition, string manipulation, hypothesis testing, and multi-step reasoning \cite{gritsevskiy2024rebus}. These benchmarks ground answers in visible content, compositional relations, symbolic clues, or domain knowledge. \cframework{} instead targets conventional creative expressions linked to images indirectly through annotated bridge paths. Complementary task views separate open recovery, candidate recognition, and explanation, while fixed targets and explicit paths support exact scoring and analysis by source and difficulty.

\subsection{Creative Cognition and Conceptual Bridging}

Cognitive accounts ground the cross-concept formulation of creative understanding. Creativity is commonly characterized by both novelty and appropriateness \cite{runco2012standard}. Creative cognition studies how retrieval, recombination, imagery, and exploration produce new but interpretable structures \cite{finke1992creative,beaty2016creative}. Associative theory links creativity to remote conceptual connections \cite{mednick1962associative}, structure mapping formalizes analogical alignment \cite{gentner1983structure}, and conceptual integration explains how meaning emerges from partial input spaces \cite{fauconnier1998conceptual}. Across these accounts, creative interpretation depends on recognizing and traversing relations between concepts that are not obviously adjacent. The \cframework{} framework operationalizes these mechanisms through a cross-concept network, where imageable substitutes and conventional target slots are connected by explicit bridge paths. Bridge count and depth translate associative distance into structured difficulty levels, while the retained paths make each hypothesized relation inspectable.

\subsection{Creative-Task Benchmarks}

Creativity benchmarks cover divergent generation, lateral problem solving, and multimodal creation. Psychometric tests compare model and human creativity, while BRAINTEASER and LatEval assess lateral thinking \cite{bellemarepepin2026divergent,jiang2023brainteaser,huang2024lateval}. Generative benchmarks evaluate literary outputs, visual humor, and image-conditioned creative responses. Related studies examine causal interventions and human--AI effects on novelty and diversity \cite{chakrabarty2024art,zhang2024humor,fang2025creationmmbench,huang2025causality,zhou2024generative,doshi2024generative}. These approaches typically rely on diversity proxies, task success, or open-ended judgments. \cset{} advances receptive creative evaluation by requiring models to recover a fixed conventional target from a creative image and connect visible clues to target slots through annotated bridge paths. Exact recovery supports reproducible scoring, while the cross-concept network distinguishes visual-recognition failures from decoding failures.

\section{Benchmark Design}

\subsection{Cross-Concept Network Construction}

The foundation of the \cframework{} framework is a manually annotated chengyu-oriented cross-concept network. We build this network once and reuse it across all synthetic encoding, rather than generating associations separately for each item. Constructing the network ahead of item generation brings two benefits. Every substitute concept in every item traces back to a human-written and reviewed associative path, and the same slot and path inventory supports systematic enumeration of item batches.
Annotation proceeds in two rounds. First, two annotators worked independently on each target chengyu. They marked one or more contiguous character spans as candidate concept slots, then associated outward from every slot and wrote one- or multi-step bridge chains toward alternative concepts. A chain may traverse phonetic, lexical, semantic, object, role, part--whole, or culturally grounded associations. Each node in one chain provides a candidate landing concept for replacing the source slot and rendering a visual clue. 
The two annotators produced separate slot schemes and chain sets for the same targets.
The second round combines internal verification with external review. The two annotators first cross-checked each other's slot boundaries and bridges, flagging spans that break morpheme structure and bridges whose steps lack a recoverable relation. A third reviewer then examined both annotation sets, adjudicated the flagged records, and revised slot boundaries and bridges where needed. We parse the reviewed records into exact chengyu-anchored spans, flatten alternative slot schemes, merge duplicate chains by target, effective slot, and node sequence, and retain source provenance for each surviving path. The resulting network contains 47 target chengyu, 168 anchored slots, and 758 deduplicated bridges. Figure~\ref{fig:overview} places this annotation stage before item-level path selection. The network is a reusable encoding resource.

\subsection{Cross-Concept Encoding}

Given the reviewed network, the encoding component turns a target chengyu into an image item with known latent structure. The process fixes a target \(y\), selects one or more non-overlapping target slots, and follows one bridge path from each selected slot to an imageable substitute concept. It then composes the substituted phrase by writing each substitute back into its slot position, and prompts an image generator with a scene description that renders the substitutes as visual clues. The image shows the substitutes. The target itself never appears. Enumerating valid combinations of target slots and bridge paths produces item batches whose cross-concept structure is known by construction, because every item records which slots were replaced, which paths were followed, and how deep each path runs.
Scene prompts are reviewed to prevent leakage of the gold chengyu, the substituted phrase as a title, or explicit task meta-language. The target stays fixed while slot choices, relation types, and bridge depths vary, which enables exact scoring and stratification by bridge count and depth across batches. The framework expands by adding reviewed target chengyu and bridge paths to the network, and every new path immediately becomes available to the encoder.

\begin{figure}[t]
\centering
\includegraphics[width=\linewidth]{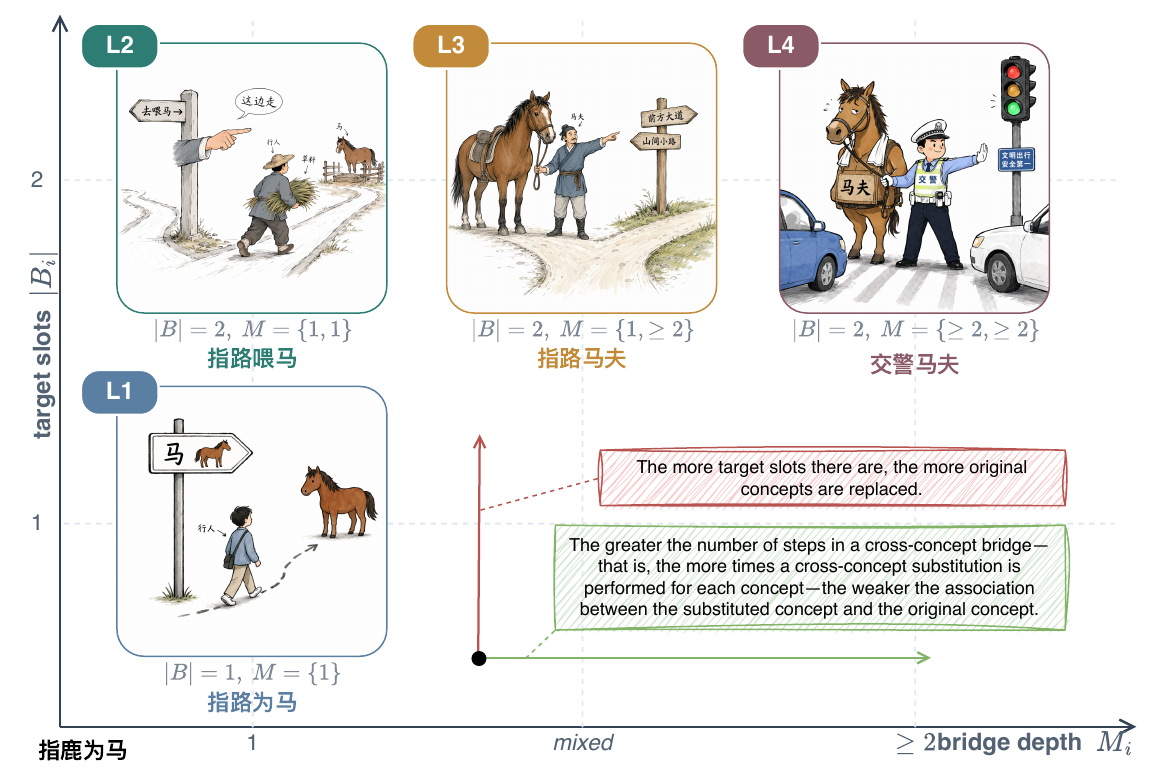}
\caption{Encoding principles for the four synthetic levels. Each realized example is placed at its own encoding coordinates, with bridge depth on the \(x\)-axis and target-slot count on the \(y\)-axis. L3 falls between the depth-1 and depth-\(\geq\)2 columns since its two paths differ. All four points instantiate the same target, \emph{zh\v{\i} l\`{u} w\'{e}i m\v{a}} (``calling a deer a horse''), under the four structures.}
\label{fig:level-construction}
\end{figure}

Synthetic difficulty follows directly from the encoding parameters. Level 1 uses one slot and one bridge step. Level 2 uses two non-overlapping slots, each with one bridge step. Level 3 uses two non-overlapping slots where one path is one-step and the other runs two or more steps. Level 4 uses two non-overlapping slots where both paths run two or more steps. Figure~\ref{fig:level-construction} places a realized example at its encoding coordinates, jointly indexed by bridge depth and target-slot count. 
The rendered content illustrates each level, while Equation~\ref{eq:difficulty} defines it from the annotated path set \(\mathcal{B}_i\) and depth multiset \(M_i\). The level scheme provides a structured proxy for cross-concept integration demand. A one-step path should usually be easier to reverse than a multi-step path, and two slots should usually require more integration than one.

\subsection{\cset{} Composition}

We instantiate the \cframework{} framework as \cset{} with synthetic and web-collected sources, summarized in Table~\ref{tab:cset-composition}. The synthetic subset is produced by the encoding component and follows four construction-defined difficulty levels. The web-collected subset consists of human-created cross-concept chengyu figures gathered from online sources. For each figure, we manually reconstruct and review its cross-concept relations, bridge paths, and reasoning process, placing both sources under the same annotation schema. The L1--L4 labels describe synthetic construction parameters. The reviewed bridge paths and reasoning processes characterize the web-collected figures, whose difficulty also reflects visual density, artistic convention, cultural context, and creator-specific encoding choices.

\begin{table}[t]
\centering
\small
\setlength{\tabcolsep}{4pt}
\begin{tabular}{@{}llr@{}}
\toprule
Category & Breakdown & Count \\
\midrule
Synthetic items & L1--L4 (47/47/46/44) & 184 \\
Web-collected figures & Online sources & 37 \\
All items & 84 target chengyu & 221 \\
Evaluation cases & T1--T5 & 1{,}105 \\
Primary cases & T1--T4 & 884 \\
\bottomrule
\end{tabular}
\caption{Composition and evaluation scale of \cset{}.}
\label{tab:cset-composition}
\end{table}

\subsection{Formal Task Definition}

Let \(\mathcal{Y}\) be the chengyu inventory and let each evaluation item be
\begin{equation}
z_i=(I_i,y_i,\mathcal{B}_i,o_i),\qquad
y_i=(c_{i1},c_{i2},c_{i3},c_{i4})\in\mathcal{Y}.
\label{eq:item}
\end{equation}
Here \(I_i\) is the image, \(y_i\) is the gold four-character chengyu, \(\mathcal{B}_i\) is the set of annotated bridge paths, and \(o_i\) records the source type, web-collected or synthetic. Written in the decoding direction, a bridge path carries a visible substitute concept back to one target slot:
\begin{equation}
b_{ij}: v_{ij0}\rightarrow v_{ij1}\rightarrow\cdots\rightarrow v_{ijm_{ij}}
\mapsto c_{is(j)} .
\label{eq:bridge}
\end{equation}
The depth \(m_{ij}\) counts bridge steps, and \(s(j)\) identifies the target slot reached by path \(j\). Encoding traverses these paths from slot to substitute, and a solver must traverse them in reverse. For synthetic items, the difficulty level is defined by bridge count and the multiset of bridge depths \(M_i=\{m_{ij}:b_{ij}\in\mathcal{B}_i\}\):
\begin{equation}
D(z_i)=
\begin{cases}
L1, & |\mathcal{B}_i|=1,\ M_i=\{1\},\\
L2, & |\mathcal{B}_i|=2,\ M_i=\{1,1\},\\
L3, & |\mathcal{B}_i|=2,\ M_i=\{1,\geq 2\},\\
L4, & |\mathcal{B}_i|=2,\ M_i=\{\geq 2,\geq 2\}.
\end{cases}
\label{eq:difficulty}
\end{equation}
Equation~\ref{eq:difficulty} assigns L1--L4 labels only to synthetic items. The web-collected figures retain their manually constructed bridge paths and reasoning processes without synthetic difficulty labels.

Cross-concept encoding must also leave the item solvable. We impose both conceptual non-identity and bridge recoverability. Let \(V(I_i)\) denote the visible concepts a solver can extract from the image. The visible concepts need not literally instantiate the target chengyu, but the reviewed annotations must license a path to it:
\begin{equation}
V(I_i)\not\Rightarrow_{\mathrm{literal}} y_i,
\qquad
V(I_i)\overset{\mathcal{B}_i}{\Rightarrow} y_i.
\label{eq:crossconcept}
\end{equation}
The first condition prevents direct target illustration from collapsing cross-concept decoding into literal recognition. The second prevents the answer from becoming arbitrary. The evaluation target remains \(y_i\), and the bridge count and depths in \(\mathcal{B}_i\) define the synthetic difficulty level. The annotations describe the intended latent relations and are not model inputs except when a task setting supplies a general bridge hint.

\subsection{Task Suite}

Each item is instantiated into five reader-facing task settings. We use T1--T5 in the paper to avoid exposing internal experimental identifiers. All five settings probe the decoding direction under different amounts of support. T1 is image-only recovery, where the model sees the image and must output the chengyu. T2 adds a general bridge hint stating that the image may express a chengyu through homophony, character decomposition, substitution, role association, object association, or related cross-concept relations. The hint names representative relation types and never reveals item-specific bridge paths. T3 gives a four-way candidate set containing the gold chengyu and three distractors drawn from the \cset{} inventory, and the model must output the chengyu string itself rather than an option letter. T4 asks for a free answer plus structured explanation in JSON. T5 gives the gold chengyu and asks only for a structured explanation.
Formally, let \(h\) be the bridge-hint text, \(C_i\) the candidate set, \(\sigma\) the requested JSON schema, and \(\hat r\) a free-text or structured rationale. The task views are
\begin{equation}
\begin{aligned}
T_1 &: I_i\mapsto \hat y, &
T_2 &: (I_i,h)\mapsto \hat y,\\
T_3 &: (I_i,C_i)\mapsto \hat y, &
T_4 &: (I_i,\sigma)\mapsto(\hat y,\hat r),\\
T_5 &: (I_i,y_i,\sigma)\mapsto \hat r.
\end{aligned}
\label{eq:taskviews}
\end{equation}
The five settings separate distinct sources of difficulty. T1 measures unconstrained decoding from the image. T2 tests whether an explicit statement of the bridge phenomenon helps. T3 measures recognition when the answer search space is collapsed to a candidate set. T4 tests whether asking for an explanation changes answer recovery and whether the model can satisfy a structured output format. T5 separates explanation quality from answer search.

\section{Experiments}

We organize the empirical study around four research questions on cross-concept decoding.
\textbf{RQ1} asks how well current MLLMs decode cross-concept chengyu items overall.
\textbf{RQ2} asks where the decoding bottleneck lies, in visual cue recognition, in relation inference, or in open answer-space search.
\textbf{RQ3} asks which encoding parameter contributes more difficulty, the number of target slots or the depth of bridge paths.
\textbf{RQ4} asks whether models reconstruct the encoded bridge paths faithfully once the answer is known, and what shared failure patterns emerge on items that defeat every model.

\subsection{Experiment Settings}

We evaluate ten MLLMs on all 221 items in \cset{} under the five task settings. The primary comparison uses the 884 answer-bearing T1--T4 cases, while T5 supplies 221 known-answer explanations for targeted analysis. The evaluated models split into two inference types. The API type covers GPT-5.5~\cite{openai2025gpt5}, Kimi-K2.6~\cite{kimiteam2026k25}, Grok-4.3, MIMO-v2.5~\cite{xiao2026mimov2flash}, and Mistral-Large-3. The Local type covers GLM-4.1V-9B-Thinking~\cite{hong2025glm41v}, InternVL3.5-8B~\cite{wang2025internvl35}, Gemma3n-E4B-it~\cite{gemmateam2025gemma3}, Nemotron-Nano-12B-v2-VL~\cite{nvidia2025nemotronnanov2vl}, and Qwen3.5-9B~\cite{bai2025qwen3vl}. Local models were served with vLLM using NVIDIA A40, L40S, A100, H100 NVL, H200, and RTX PRO 6000 Blackwell GPUs.
We report results by task setting, source, and synthetic construction level. T5 is analyzed separately because its prompt contains the target answer.


The current metric is strict exact answer match after lightweight normalization. Let \(\nu(\cdot)\) be the normalization function and let \(\mathcal{A}_i\) contain the accepted gold strings for item \(i\).
Each \(\mathcal{A}_i\) contains exactly the canonical chengyu string and alias sets are supported by the scorer but unused. For model \(m\), task \(t\), and predicted answer \(\hat y_{i,t}^{(m)}\), exact recovery is
\begin{equation}
e_m(i,t)=
\mathbf{1}\left[\nu\!\left(\hat y_{i,t}^{(m)}\right)\in
\{\nu(a):a\in\mathcal{A}_i\}\right].
\label{eq:exact}
\end{equation}
The primary leaderboard score is
\begin{equation}
\mathrm{Primary}(m)=
\frac{1}{4N}\sum_{i=1}^{N}\sum_{t=1}^{4} e_m(i,t),
\qquad N=221.
\label{eq:primary}
\end{equation}
T5 is excluded from Equation~\ref{eq:primary} because its prompt supplies the gold chengyu. For T5, we compare the generated rationale with the annotated visible cues, bridge paths, and target-fragment alignment rather than treating answer recovery as evidence of understanding.

\subsection{RQ1: Overall Decoding Performance}

RQ1 asks how well current MLLMs decode cross-concept chengyu items. Table~\ref{tab:leaderboard} shows the completed official leaderboard together with the task-form breakdown. Closed models and open models separate into two clear tiers, and the top score of 50.7\% sits far above the best open score of 18.1\%.
\begin{table}[t]
\centering
{\small
\setlength{\tabcolsep}{4pt}
\begin{tabular}{@{}lccccc@{}}
\toprule
Model & T1 & T2 & T3 & T4 & Primary \\
\midrule
GPT-5.5 & 35.7 & 40.3 & 87.3 & 39.4 & \textbf{50.7} \\
Kimi-K2.6 & 32.6 & 39.4 & 85.5 & 34.4 & 48.0 \\
Grok-4.3 & 23.5 & 24.4 & 75.1 & 24.0 & 36.8 \\
MIMO-v2.5 & 16.7 & 19.0 & 67.0 & 21.3 & 31.0 \\
GLM-4.1V-9B & 5.0 & 7.2 & 53.4 & 6.8 & 18.1 \\
InternVL3.5-8B & 5.4 & 5.0 & 52.5 & 8.1 & 17.8 \\
Gemma3n-E4B-it & 2.3 & 2.3 & 58.4 & 2.7 & 16.4 \\
Mistral-Large-3 & 2.7 & 3.6 & 42.1 & 5.4 & 13.5 \\
Nemotron-Nano-12B & 1.4 & 0.9 & 46.2 & 1.4 & 12.4 \\
Qwen3.5-9B & 6.3 & 4.1 & 22.2 & 4.1 & 9.2 \\
\bottomrule
\end{tabular}
}
\caption{Primary accuracy and task-form breakdown.}
\label{tab:leaderboard}
\end{table}
The results show that cross-concept decoding remains far from solved. Even the top model misses roughly half of the primary cases. The two strongest closed models track each other closely in aggregate accuracy, yet they diverge by several points once individual task views are compared, so overall parity can hide task-specific differences. The best open model sits closer to the weakest closed model than to the top of the closed tier. Scale and training recipe therefore dominate the current ranking, yet no recipe brings a model near the ceiling.

\subsection{RQ2: Locating the Decoding Bottleneck}

\begin{figure}[t]
\centering
\includegraphics[width=\linewidth]{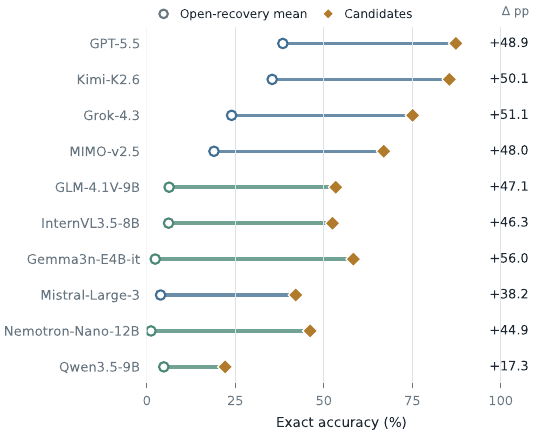}
\caption{Candidate constraint lift, measured as candidate-recognition accuracy minus the mean of the three open recovery settings. The lift is large for most models, showing that answer-space search is a central bottleneck.}
\label{fig:lift}
\end{figure}

RQ2 asks where cross-concept decoding fails. Three stages could be responsible: missing visual cues, failing to infer the latent relations from recognized cues, or failing to search the open answer space once usable relations are available. The task suite separates these stages because T3 removes the search burden while keeping perception and relation inference unchanged.
Table~\ref{tab:leaderboard} also shows the task-form profiles. The candidate constraint produces the most striking effect, lifting accuracy far above the open settings for every evaluated model, including the weakest ones.
Figure~\ref{fig:lift} summarizes this gap. Across the evaluated models, candidate recognition exceeds the mean of T1, T2, and T4 by 17.3 to 56.0 percentage points. The image therefore supplies enough evidence for partial alignment once the target enters a small candidate set, even when that evidence cannot drive unconstrained retrieval. Visual perception is not the limiting stage for most items, since invisible cues could not let a collapsed answer space raise accuracy by tens of points.

The bridge hint isolates the relation-inference stage rather than the search stage. T2 names possible relation types without identifying the relevant cue, substituted concept, or target slot, and it produces much smaller gains than candidates. T4 behaves like the other open settings rather than approaching candidate-level accuracy.
Item-level pairing sharpens the picture. Table~\ref{tab:paired-views} pairs T1 and T4 for every item and separately counts candidate-only recovery. Requiring an explanation changes which items are solved rather than producing a uniform shift. Most models recover more items only in T4 than only in T1, though the direction reverses for the weakest model. Explanation prompting can therefore help on some items and hurt on others, even with a fixed image and answer space.
\begin{table}[t]
\centering
{\scriptsize
\setlength{\tabcolsep}{3.2pt}
\begin{tabular}{@{}lrrr@{}}
\toprule
Model & T1 only & T4 only & Candidate only \\
\midrule
GPT-5.5 & 10 & 18 & 89 \\
Kimi-K2.6 & 16 & 20 & 90 \\
Grok-4.3 & 6 & 7 & 99 \\
MIMO-v2.5 & 6 & 16 & 90 \\
GLM-4.1V-9B & 1 & 5 & 104 \\
InternVL3.5-8B & 3 & 9 & 96 \\
Gemma3n-E4B-it & 1 & 2 & 122 \\
Mistral-Large-3 & 0 & 6 & 83 \\
Nemotron-Nano-12B & 0 & 0 & 99 \\
Qwen3.5-9B & 7 & 2 & 37 \\
\bottomrule
\end{tabular}
}
\caption{Paired outcomes over 221 items. ``T1 only'' and ``T4 only'' count disagreements between image-only and free-explanation recovery. ``Candidate only'' counts T3 successes for which all three open settings, T1, T2, and T4, fail.}
\label{tab:paired-views}
\end{table}
Candidate-only recovery quantifies the search bottleneck directly. Roughly half to the majority of the closed models' T3 successes occur on items missed by all three open settings, and the share rises further for most open models, reaching at least three quarters of all T3 successes. The answer to RQ2 follows from these contrasts. Visual cue recognition is largely intact, relation inference is partially available and surfaces once the search space collapses, and open answer-space search is the dominant bottleneck across all evaluated models.

\subsection{RQ3: Slots versus Bridge Depth}

\begin{figure}[t]
\centering
\includegraphics[width=0.95\linewidth]{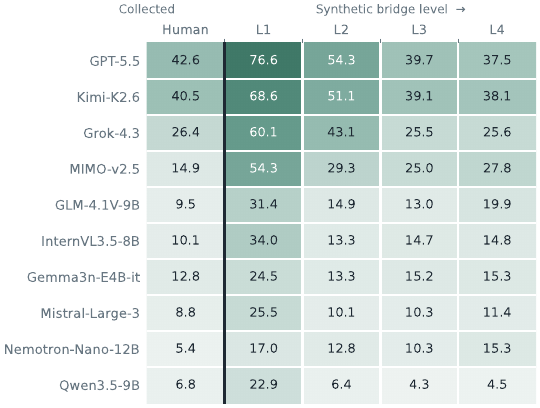}
\caption{Difficulty profile across source and synthetic levels.}
\label{fig:difficulty}
\end{figure}

RQ3 asks which encoding parameter contributes more difficulty. The level design permits a direct comparison because adjacent levels change exactly one parameter. Moving from L1 to L2 adds a second slot while keeping every path one-step. Moving from L2 to L3 deepens one path while keeping two slots. Moving from L3 to L4 deepens the second path.
Figure~\ref{fig:difficulty} shows performance by source and synthetic level. Accuracy falls sharply from L1 to L2, drops more moderately from L2 to L3, and shows no further decline from L3 to L4, reversing slightly for several models.

The answer to RQ3 is that slot count dominates. Averaged across all ten models, the L1-to-L2 slot transition produces roughly three times the mean accuracy drop of the L2-to-L3 depth transition, and additional depth beyond the first multi-step path yields no further measurable difficulty. The level buckets contain different targets and rendered images, so these adjacent comparisons describe associations with the annotated encoding rather than controlled causal effects.
The web-collected subset stays difficult for all models, scoring below the synthetic average for every model. Human creators combine dense composition, cultural context, and free-form encoding choices, so this subset checks that the synthetic difficulty scheme does not exhaust the phenomenon.

\subsection{RQ4: Explanation Fidelity and Shared Failures}

RQ4 asks whether models reconstruct the encoded bridge paths faithfully and what items defeat every model. We address the first part with a paired T4--T5 case study and the second with a corpus-level bad-case analysis.

\begin{figure}[t]
\centering
\includegraphics[width=\linewidth]{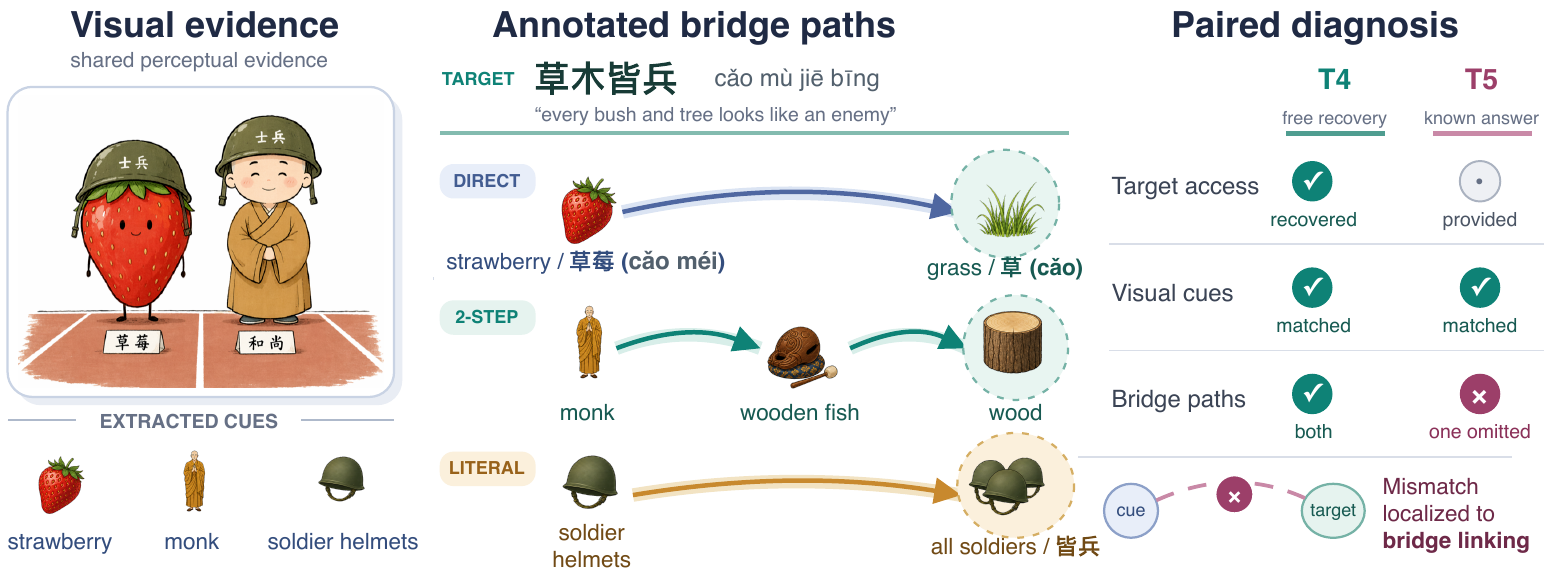}
\caption{Paired T4--T5 diagnosis for \emph{c\v{a}o m\`{u} ji\={e} b\={\i}ng} (``every bush and tree looks like an enemy''). Both settings recover the same visual cues, while only T4 reconstructs both annotated bridge paths.}
\label{fig:t5case}
\end{figure}

\paragraph{Known-answer explanation.}
T5 isolates explanation behavior after answer retrieval has been removed. 
We analyze each explanation in two stages. First, we compare the reported perceptual cues with the annotated visual elements. Second, we compare the proposed cue-to-target links with the gold bridge paths and their aligned target fragments. T4 on the same image provides an independent cross-check because both its answer and rationale are generated without access to the target. Figure~\ref{fig:t5case} aligns the shared visual evidence with the annotated L3 paths and the paired model diagnoses.
For GPT-5.5, the paired outputs agree at the perceptual layer but diverge at the bridge layer. In visual recognition, T4 identifies the labeled strawberry, the labeled monk, and the soldier helmets, and T5 reports the same three elements, ruling out object recognition as the bottleneck here. In bridge linking, T4 correctly predicts the target and explicitly reconstructs both annotated paths, including the culturally mediated association from monk to wooden fish and then to wood. T5 instead maps the strawberry directly to ``grass and wood,'' omits the monk path, and explains the scene through the conventional meaning that vegetation is being treated as soldiers. The explanation stays fluent but does not recover the encoding that generated the image. The correct T4 prediction proves that the image contains sufficient evidence for both target recovery and the intended paths, so the T5 mismatch arises at the decoding step that links a recognized entity through an intermediate cultural concept to its slot. A correct answer field or valid JSON alone does not establish explanatory faithfulness.

\paragraph{Bad cases shared across models.}
We examine items that resist every model. Across the ten evaluated models, 83 of the 221 items are never solved by any model in any of the three open settings, and 6 items remain unsolved even when candidates are provided. The open-unsolved items concentrate in the higher synthetic levels, matching the RQ3 finding that two-slot encoding drives shared failure.
One fully unsolved item makes the shared failure mode concrete. The L3 item encodes \emph{y\={\i} y\`{e} zh\`{a}ng m\`{u}} (``a leaf blocks one's view'') as the phrase ``one coconut mother-in-law,'' following a one-step near-homophone path from leaf to coconut and a two-step path from block-the-eye through a homophone to mother-in-law. The rendered image shows a mother-in-law at a dinner table with a coconut and straw at hand. Model outputs reveal decoding anchored on the surface. GPT-5.5's structured rationale correctly lists the mother-in-law label, the coconut, and the straw, then assembles them literally into \emph{yu\`{e} m\v{u} c\`{\i} z\`{\i}} (``General Yue's mother tattoos characters on his back''), treating the visible concepts as target fragments instead of reversing their bridges. Under candidates, every strong model selects the distractor \emph{y\`{e} g\={o}ng h\`{a}o l\'{o}ng} (``Lord Ye loves loongs''), which shares the leaf character with the gold target, rather than the target itself. All ten models solve none of the four answer-bearing tasks on this item, yet the strongest models reconstruct both encoded paths in T5 once the answer is given. The failure is therefore a decoding failure rather than a perceptual or knowledge gap, since visible substitutes are consumed literally instead of being traversed backward through their encoded relations.


\section{Conclusion}

We introduce \cframework{}, a cognition-inspired framework that makes receptive creative understanding measurable through cross-concept encoding and decoding. We instantiate the framework as \cset{}, combining synthetic items with human-created figures collected from online sources under a shared annotation schema. \cframework{} builds on a manually annotated chengyu-oriented cross-concept network and integrates structured item construction, task instantiation, answer extraction, exact scoring, and stratified analysis. Across ten MLLMs, even the strongest model recovers only about half of the primary cases, leaving the leaderboard far from saturated. Candidate constraints produce large gains, while open models remain far behind the strongest closed systems. Further analysis identifies open answer-space search as the dominant bottleneck, shows that slot count influences difficulty more strongly than bridge depth, and finds that models often consume visible substitutes literally rather than decode their cross-concept relations. These findings demonstrate substantial headroom in cross-concept understanding and establish \cframework{} as a structured, extensible foundation for evaluating how MLLMs decode creatively encoded meaning across concepts.

\bibliography{chengyu_bench}

\appendix
\setcounter{table}{0}
\renewcommand{\thetable}{S\arabic{table}}
\setcounter{figure}{0}
\renewcommand{\thefigure}{S\arabic{figure}}

\section{Theoretical Scope and Operational Assumptions}

The theoretical contribution of \cframework{} is a cognition-grounded formulation of receptive creative understanding for MLLMs. It connects three levels of description. Cognitive theories motivate cross-concept connection as a core mechanism of creative interpretation. The encoding and decoding formulation identifies the two directions of the process. The cross-concept network then makes the formulation operational through target slots, bridge paths, landing concepts, and recoverable answers.

The evaluation rests on four explicit assumptions. First, an item has a conventional target whose written form is fixed. Second, the visible concepts differ from a literal depiction of that target. Third, every intended substitution has a reviewed bridge path back to a target slot. Fourth, the annotated path records the intended construction but does not exclude other plausible interpretations of the same image. Exact recovery measures whether a model reaches the intended conventional target. The L1--L4 levels describe construction parameters. They do not assert an intrinsic psychological scale independent of targets and images.

The formal objects in the main paper define an item, a bridge path, the four synthetic levels, the recoverability condition, the five task views, and the primary score. The experiments test the resulting predictions at the benchmark level. Candidate constraints isolate answer-space search, paired T1 and T4 outcomes measure prompt-form sensitivity, level comparisons separate slot count from bridge depth, and T5 enables direct inspection of visual recognition and bridge reconstruction.

\section{Data and Resource Release}

Upon publication, we will release the complete \cset{} evaluation set, the reviewed cross-concept annotations, all synthetic and collected images, task instances, model outputs used in the paper, and the code required to construct and evaluate the benchmark. Table~\ref{tab:supp-artifacts} summarizes these resources. We will license the code under Apache-2.0 and the data, annotations, prompts, and benchmark metadata under CC BY-NC 4.0. The evaluation package will include the collected figures.

\begin{table}[H]
\centering
\small
\setlength{\tabcolsep}{5pt}
\begin{tabular}{@{}>{\raggedright\arraybackslash}p{0.22\textwidth}>{\raggedright\arraybackslash}p{0.38\textwidth}>{\raggedright\arraybackslash}p{0.32\textwidth}@{}}
\toprule
Resource & Contents & Reproducibility role \\
\midrule
Cross-concept annotations & Reviewed target slots, bridge chains, landing concepts, relation types, and annotation provenance & Supports inspection and reuse of the encoding resource \\
\cset{} items & Images, canonical answers, source labels, difficulty levels, and aligned bridge annotations & Reconstructs the evaluated item collection \\
Task instances & T1--T5 prompts, candidate sets, expected answers, and explanation annotations & Reproduces the five evaluation settings \\
Model predictions & Original responses and scored predictions for every reported model & Recomputes task scores and stratified analyses \\
Construction and evaluation code & Network processing, item construction, task instantiation, scoring, and analysis procedures & Reproduces benchmark construction and reported results \\
\bottomrule
\end{tabular}
\caption{Resources included in the publication release.}
\label{tab:supp-artifacts}
\end{table}

\subsection{Data Card}

\cset{} contains 221 items covering 84 target chengyu. The synthetic subset contains 47 L1, 47 L2, 46 L3, and 44 L4 items. The collected subset contains 37 human-created figures. Each item has five task views. This produces 1,105 evaluation cases, of which 884 answer-bearing T1--T4 cases determine the primary score.

Each item includes a source type, image, canonical answer, task prompt, and accepted-answer list. Synthetic items additionally include the selected target slots, complete bridge chains, landing concepts, substituted phrase, generation prompt, and L1--L4 label. Collected items include visible cues, bridge units, target-fragment alignment, and a reviewed rationale. The benchmark contains no personal attributes or private user records.

\section{Cross-Concept Network Annotation}

Two annotators independently processed each of the 47 synthetic target chengyu. Each annotator selected one or more contiguous character spans that could function as concept slots. They then associated outward from each slot and wrote one-step or multi-step chains that ended at an imageable concept. The two annotation sets were retained separately during the first round.

In the second round, the annotators cross-checked slot boundaries and bridge steps. They flagged spans that broke the intended lexical unit and steps that lacked a recoverable relation. A third reviewer examined both annotation sets, adjudicated flagged records, and revised slots and chains. The reviewed annotations were normalized to exact character spans. Alternative slot schemes were retained, duplicate chains were merged by target, effective slot, and node sequence, and source provenance was preserved. The final resource contains 47 targets, 168 anchored slots, and 758 deduplicated bridge chains.

\begin{table}[H]
\centering
\small
\begin{tabular}{@{}lp{0.65\linewidth}@{}}
\toprule
Relation & Annotation criterion \\
\midrule
Phonetic & Homophony or near homophony links two spoken forms \\
Lexical & A character, word, or conventional expression licenses the next node \\
Semantic & The next node is connected by meaning, category, or functional similarity \\
Object & An object conventionally evokes another object or its use \\
Role & An agent, profession, or social role evokes a related concept \\
Part--whole & A salient component links to an object, event, or category \\
Cultural & Shared stories, customs, named entities, or conventional knowledge license the step \\
\bottomrule
\end{tabular}
\caption{Relation families used during bridge annotation. A chain may combine several families.}
\label{tab:supp-relations}
\end{table}

Every retained chain must satisfy slot alignment, conceptual non-identity, stepwise recoverability, and imageability of its landing concept. Slot alignment requires the chain to terminate at the annotated target span. Conceptual non-identity prevents a landing concept from simply repeating the target. Stepwise recoverability requires each adjacent pair to have an interpretable relation. Imageability requires the final node to support a concrete visual scene.

\section{Synthetic Construction and Quality Control}

The encoder enumerates non-overlapping target slots and reviewed bridge chains. L1 selects one one-step chain. L2 selects two one-step chains. L3 selects one one-step chain and one chain of depth two or greater. L4 selects two chains of depth two or greater. The build retains at most one item for each target and level. It removes unchanged substitutions, overlapping slots, malformed phrases, and duplicate items. This process yields 184 items from 188 possible target-level cells.

All scene descriptions were manually reviewed. Review removed direct appearances of the gold chengyu, titles that repeated the substituted phrase, and explicit task instructions. The final 184 images were generated with GPT Image 2 at a resolution of 1024 by 1024 pixels. The release includes each final image together with its target, level, substituted phrase, scene description, prompt, and gold bridge paths. Item selection, candidate construction, and task instantiation are deterministic once the reviewed annotations are fixed.

\section{Task Instantiation and Scoring}

Table~\ref{tab:supp-tasks} summarizes the five task settings. All task prompts are written in Chinese to match the language of the target expressions. T3 distractors are selected from the 84-answer inventory. Candidates that share more Chinese characters with the gold target are ranked first, and ties are resolved deterministically.

\begin{table}[H]
\centering
\small
\setlength{\tabcolsep}{4pt}
\begin{tabular}{@{}clp{0.57\textwidth}@{}}
\toprule
Task & Required output & Information supplied \\
\midrule
T1 & Chengyu & Image only \\
T2 & Chengyu & Image and a general cross-concept hint \\
T3 & Chengyu & Image and four candidate chengyu \\
T4 & Chengyu and rationale & Image and the structured explanation schema \\
T5 & Rationale & Image, gold chengyu, and the structured explanation schema \\
\bottomrule
\end{tabular}
\caption{Required outputs and supplied information for the five task settings.}
\label{tab:supp-tasks}
\end{table}

T1--T3 score one predicted chengyu, while T4 scores the answer supplied with the structured rationale. Normalization removes whitespace and common ASCII and Chinese punctuation. The accepted-answer set contains only the canonical four-character target. T5 is excluded from primary accuracy because its prompt supplies the target.

GLM-4.1V-9B-Thinking emits a visible reasoning segment followed by an explicit final-answer payload, which supplies its scored prediction. Qwen3.5-9B is scored only when it produces a final answer. Responses that terminate before that answer remain incorrect. The publication release includes the original responses and final scored predictions.

T5 explanations are analyzed in two stages. The first stage compares reported perceptual cues with the annotated visual elements. The second compares cue-to-target links with the annotated bridge chains and target fragments. T4 on the same item provides an answer-hidden cross-check.

\section{Experimental Configuration}

Each official row is one complete pass over all 1,105 task cases. Temperature is set to zero when the endpoint exposes that parameter. No output-token cap is applied. API models use their provider-managed serving context and documented default interaction mode. Local inference runs on Linux with Python 3.11.11, CUDA 13.0.2, and vLLM 0.23.0. Each local process uses one listed NVIDIA accelerator with 40 to 141 GB of GPU memory.

\begin{table}[H]
\centering
\scriptsize
\setlength{\tabcolsep}{3.3pt}
\begin{tabular}{@{}>{\raggedright\arraybackslash}p{0.19\textwidth}>{\raggedright\arraybackslash}p{0.13\textwidth}>{\raggedright\arraybackslash}p{0.13\textwidth}>{\raggedright\arraybackslash}p{0.18\textwidth}>{\raggedright\arraybackslash}p{0.27\textwidth}@{}}
\toprule
Model & Interface & Context & Hardware & Final decoding and scoring setting \\
\midrule
GPT-5.5 & Provider API & Provider managed & Provider managed & Temperature omitted, no output cap, exact scoring \\
Kimi-K2.6 & Provider API & Provider managed & Provider managed & Temperature 0, no output cap, exact scoring \\
Grok-4.3 & Provider API & Provider managed & Provider managed & Temperature 0, no output cap, exact scoring \\
MIMO-v2.5 & Provider API & Provider managed & Provider managed & Temperature 0, no output cap, exact scoring \\
Mistral-Large-3 & Provider API & Provider managed & Provider managed & Temperature 0, no output cap, exact scoring \\
GLM-4.1V-9B-Thinking & vLLM & 65,536 & H100 NVL & Temperature 0, no output cap, final-answer extraction \\
InternVL3.5-8B & vLLM & 8,192 & A100 & Temperature 0, no output cap, native chat behavior \\
Gemma3n-E4B-it & vLLM & 32,768 & A100 & Temperature 0, no output cap, instruction-tuned image-first template \\
Nemotron-Nano-12B-v2-VL & vLLM & 8,192 & L40S & Temperature 0, no output cap, native chat behavior \\
Qwen3.5-9B & vLLM & 262,144 & A40, L40S, A100, H100 NVL, H200, RTX PRO 6000 Blackwell & Temperature 0, no output cap, strict final-answer scoring \\
\bottomrule
\end{tabular}
\caption{Final inference conditions. API hardware is managed by the corresponding provider.}
\label{tab:supp-config}
\end{table}

The task prompts, candidate sets, normalization, and scoring rules are fixed across models. Each reported result is computed from one complete evaluation pass.

\section{Extended Results}

Table~\ref{tab:supp-complete-results} reports the task-level numerators used to compute the main leaderboard. Every task has 221 cases. Primary accuracy divides the sum of T1--T4 correct counts by 884.

\begin{table}[H]
\centering
\small
\setlength{\tabcolsep}{4pt}
\begin{tabular}{@{}lrrrrrr@{}}
\toprule
Model & T1 & T2 & T3 & T4 & Primary & T5 \\
\midrule
GPT-5.5 & 79 & 89 & 193 & 87 & 448 & 221 \\
Kimi-K2.6 & 72 & 87 & 189 & 76 & 424 & 221 \\
Grok-4.3 & 52 & 54 & 166 & 53 & 325 & 221 \\
MIMO-v2.5 & 37 & 42 & 148 & 47 & 274 & 216 \\
GLM-4.1V-9B-Thinking & 11 & 16 & 118 & 15 & 160 & 221 \\
InternVL3.5-8B & 12 & 11 & 116 & 18 & 157 & 221 \\
Gemma3n-E4B-it & 5 & 5 & 129 & 6 & 145 & 221 \\
Mistral-Large-3 & 6 & 8 & 93 & 12 & 119 & 221 \\
Nemotron-Nano-12B-v2-VL & 3 & 2 & 102 & 3 & 110 & 218 \\
Qwen3.5-9B & 14 & 9 & 49 & 9 & 81 & 152 \\
\bottomrule
\end{tabular}
\caption{Correct-answer counts. T5 reports exact recovery of its supplied answer field and is not included in Primary.}
\label{tab:supp-complete-results}
\end{table}

The source and level breakdowns use the same four answer-bearing tasks. The 37 collected items contribute 148 primary cases. L1 and L2 each contribute 188 cases, L3 contributes 184, and L4 contributes 176. The publication release includes item-level predictions, allowing every percentage and paired count in the paper to be recomputed.

\subsection{Additional Construction Cases}

The synthetic target \emph{zh\v{\i} l\`u w\'ei m\v{a}} (``call a deer a horse'') illustrates how one target supports all four construction levels. L1 replaces \emph{zh\v{\i} l\`u} (``point at a deer'') with its homophone meaning ``give directions.'' L2 adds the one-step homophonic path from ``be a horse'' to ``feed a horse.'' L3 extends the second path from feeding a horse to a horse keeper. L4 also extends the first path from giving directions to a traffic officer. The four images therefore vary slot count and bridge depth while keeping the target fixed.

The collected case \emph{m\'ao s\`e d\`un k\={a}i} (``become suddenly enlightened'') uses two culturally grounded paths. A blocked toilet evokes the target fragment \emph{m\'ao s\`e} through its spoken description. Newton opening the blockage evokes \emph{d\`un k\={a}i} through Newton's transliterated Chinese name and the opening action. The annotation separates recognition of the toilet, Newton, and the opening action from reconstruction of the two target-fragment links.

\section{Chengyu Reference}

Tables~\ref{tab:chengyu-1}--\ref{tab:chengyu-6} list every target in \cset{}. Pinyin is italicized. The English text gives the conventional meaning used for reviewer orientation.

\begin{table}[H]
\centering
\includegraphics[page=1,width=\textwidth]{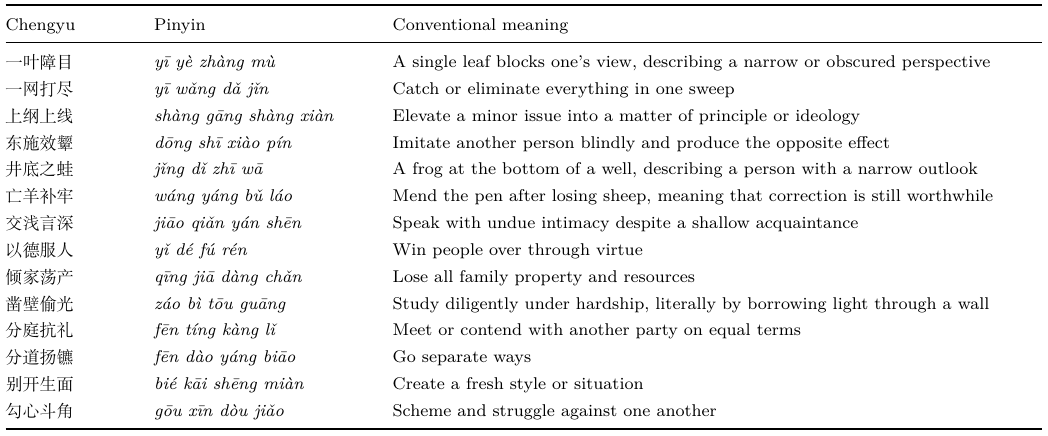}
\caption{C4-Eval chengyu reference, part 1 of 6.}
\label{tab:chengyu-1}
\end{table}

\begin{table}[H]
\centering
\includegraphics[page=2,width=\textwidth]{Figures/chengyu_reference_tables_cropped.pdf}
\caption{C4-Eval chengyu reference, part 2 of 6.}
\label{tab:chengyu-2}
\end{table}

\begin{table}[H]
\centering
\includegraphics[page=3,width=\textwidth]{Figures/chengyu_reference_tables_cropped.pdf}
\caption{C4-Eval chengyu reference, part 3 of 6.}
\label{tab:chengyu-3}
\end{table}

\begin{table}[H]
\centering
\includegraphics[page=4,width=\textwidth]{Figures/chengyu_reference_tables_cropped.pdf}
\caption{C4-Eval chengyu reference, part 4 of 6.}
\label{tab:chengyu-4}
\end{table}

\begin{table}[H]
\centering
\includegraphics[page=5,width=\textwidth]{Figures/chengyu_reference_tables_cropped.pdf}
\caption{C4-Eval chengyu reference, part 5 of 6.}
\label{tab:chengyu-5}
\end{table}

\begin{table}[H]
\centering
\includegraphics[page=6,width=\textwidth]{Figures/chengyu_reference_tables_cropped.pdf}
\caption{C4-Eval chengyu reference, part 6 of 6.}
\label{tab:chengyu-6}
\end{table}

\end{document}